\documentclass[11pt,a4paper]{article}
\usepackage[hyperref]{acl2021}
\aclfinalcopy
\def\aclpaperid{1595}

\usepackage{times}
\usepackage{latexsym}

\usepackage[T1]{fontenc}
\usepackage[utf8]{inputenc}

\usepackage{microtype}

\usepackage{times}
\usepackage{epsfig}
\usepackage{graphicx}
\usepackage{amsmath}
\usepackage{amssymb}
\usepackage{textcomp}
\usepackage{gensymb}

\usepackage{url}            % simple URL typesetting
\usepackage{booktabs}       % professional-quality tables
\usepackage{nicefrac}       % compact symbols for 1/2, etc.
\usepackage{microtype}      % microtypography
\usepackage{color}
\usepackage{amsmath}
\usepackage{amssymb}
\usepackage{amsthm}
\usepackage{mathdots}

\usepackage{bm}
\usepackage{hhline}
\usepackage{gensymb}
\newcommand*{\email}[1]{\texttt{#1}}
\usepackage{color, colortbl}
\definecolor{Gray}{gray}{0.9}

\newcommand{\myparagraph}[1]{\paragraph{#1}}
\title{Unsupervised Out-of-Domain Detection via Pre-trained Transformers \\}

\author{Keyang Xu\textsuperscript{\rm 1}, Tongzheng Ren\textsuperscript{\rm 2}, Shikun Zhang\textsuperscript{\rm 3}, Yihao Feng\textsuperscript{\rm 2}, Caiming Xiong\textsuperscript{\rm 4}\\ 
	\textsuperscript{\rm 1} Columbia University  
	\textsuperscript{\rm 2} University of Texas at Austin \\ 
	\textsuperscript{\rm 3} Carnegie Mellon University  \textsuperscript{\rm 4} Salesforce Research \\ 
	\email{kx2155@columbia.edu} ~~~\email{shikunz@cs.cmu.edu}\\
	  \email{\{tzren, yihao\}@cs.utexas.edu} 
	 ~~~\email{cxiong@salesforce.com}
}

\begin{document}
\maketitle
\begin{abstract}

Deployed real-world machine learning applications are often subject to uncontrolled and even potentially malicious inputs. Those out-of-domain inputs can lead to unpredictable outputs and sometimes catastrophic safety issues. 
Prior studies on out-of-domain detection require in-domain task labels and are limited to supervised classification scenarios.
Our work tackles the problem of detecting out-of-domain samples with only unsupervised in-domain data. 
We utilize the latent representations of pre-trained transformers and propose a simple yet effective method to transform features across all layers to construct out-of-domain detectors efficiently. 
Two domain-specific fine-tuning approaches are further proposed to boost detection accuracy.
Our empirical evaluations of related methods on two datasets validate that our method greatly improves out-of-domain detection ability in a more general scenario.\footnote{Code is available at \url{https://github.com/rivercold/BERT-unsupervised-OOD}.} 

\end{abstract}

\section{Introduction}
Although deep neural networks achieve good performance on many challenging tasks, they can make overconfident predictions for irrelevant and out-of-domain (OOD) inputs, leading to significant AI safety issues~\cite{hendrycks2016baseline}. 
Detecting out-of-domain inputs is a fundamental task for trustworthy AI applications in real-world use cases, because those applications are often subject to ill-defined queries or even potentially malicious inputs. 
Prior work on out-of-domain detection \citep[e.g.,][]{hendrycks2016baseline,lee2018simple,liang2018enhancing,hendrycks2018deep,hendrycks2020pretrained, xu2020deep} mostly requires in-domain task labels, limiting its usage to supervised classification.
However, deployed applications rarely receive controlled inputs and are susceptible to an ever-evolving set of user inputs that are scarcely labeled. 
For example, for many non-classification tasks, such as summarization or topic modeling, there are no available classifiers or task labels, which limits the practical usage of recently proposed out-of-domain detection methods.
Therefore, it is natural to ask the following question:

\emph{Can we detect out-of-domain samples using only unsupervised data without any in-domain labels?}

We regard the out-of-domain detection problem as checking whether the given test samples are drawn from the same distribution that generates the in-domain samples, which requires a weaker assumption than prior work \citep[e.g.,][]{lee2018simple, hendrycks2020pretrained}.
We suppose that there are only in-domain samples, which allows us to understand the properties of data itself regardless of tasks. Therefore, methods developed for this problem are more applicable than task-specific ones and can be further adapted to tasks where no classification labels are present, such as active learning or transfer learning.

To solve the problem, we utilize the latent embeddings of pre-trained transformers~\citep[e.g.,][]{vaswani2017attention,devlin-etal-2019-bert,liu2019roberta} to represent the input data, which allow us to apply classical OOD detection methods such as one-class support vector machines~\citep{scholkopf2001estimating}  or support vector data description \citep{tax2004support} on them.

However, the best practice on how to extract features from BERT is usually task-specific. For supervised classification, we can represent the text sequence using the hidden state of [CLS] token from the top layer. Meanwhile BERT’s intermediate layers also capture rich linguistic information that may outperform the top layer for specific NLP tasks.  By performing probing tasks on each layer, \citet{Jawahar2019WhatDB} suggest bottom layers of BERT capture more surface features, middle layers focus more on syntax and semantic features are well represented by top ones. 

As no prior knowledge about OOD samples is usually provided in practice, deciding which layer of features is the most effective for OOD detection is itself non-trivial. Some OOD samples may just contain a few out-of-vocabulary words; while others are OOD due to their syntax or semantics. 

Based on the observations above, this paper studies how to leverage all-layer features from a pre-trained transformer for OOD detection in an unsupervised manner. Our contributions are three-fold:

    \vspace{.8em}
    
    $\bullet$ 
    By analyzing all layers of (Ro)BERT(a) models, we empirically validate that it is hard to extract features from a certain layer that work well for any OOD datasets. 
    
    \vspace{.8em}  
    
    $\bullet$ We propose a computationally efficient way to transform all-layer features of a pre-trained transformer into a low-dimension one. We empirically validate that the proposed method outperforms baselines that use one-layer features or by simple aggregations of all layers.
    
    \vspace{.8em}
    
    $\bullet$ We propose two different techniques for fine-tuning a pre-trained transformer to further improve its capability of detecting OOD data.

\begin{figure*}[t]
    \centering
    \includegraphics[width=\textwidth]{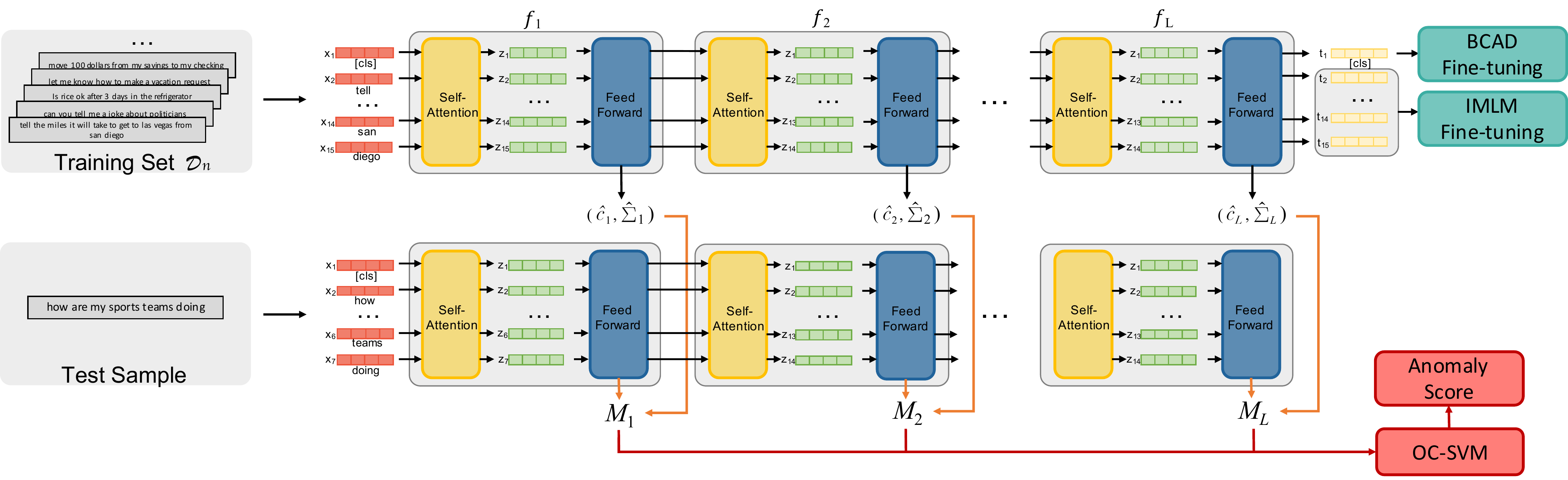}  \\
    \caption{An overview of using Mahalanobis distance features (MDF) extracted from a pre-trained transformer $f$ to detect out-of-domain data. We estimate mean $\hat{c}_l$ and covariance matrix $\hat{\Sigma}_l$ for each layer of $f$ by samples from an unsupervised training set $\mathcal{D}_n$; and then extract MDF of $\mathcal{D}_n$ to optimize a OC-SVM. Given an unseen test sample, its feature $M$ is extracted using $\hat{c}_l$ and $\hat{\Sigma}_l$ and then fed into OC-SVM for an anomaly score. Two domain-specific fine-tuning methods, IMLM and BCAD, can be further applied to BERT to boost detection accuracy.}
    \label{fig:overview}
    \vspace{-0.5em}
\end{figure*}

\section{Problem Setup}

\label{sec:setup}

Assume that we have a collection of text inputs $\mathcal{D}_{n} := \{\bm{x}_i\}_{i=1}^{n}$, we want to construct an out-of-domain detector that takes an unseen new input $\bm{u}$ and determines whether $\pmb{u}$ comes from the same distribution that generates $\mathcal{D}_{n}$. 
We adopt a more practical setting where we have \textbf{no prior knowledge} of what out-of-domain inputs look like. In this case, training a domain classifier directly is not feasible.
The out-of-domain  detector can be described mathematically as:
$$
    g(\pmb{u}, \epsilon) = \begin{cases}\rm{\textbf{True}} & \rm{if} ~~\mathcal{I}(\pmb{u}) \leq  \epsilon\,, \\ \rm{\textbf{False}} & \rm{if} ~~\mathcal{I}(\pmb{u}) > \epsilon\,, \end{cases}
$$
where $\mathcal{I}(\cdot)$ denotes the anomaly score function, and $\epsilon$ is a chosen threshold to ensure that the true positive (in-domain) rate is at a certain level (e.g., $95\%$) ~\citep{hendrycks2016baseline,liang2018enhancing,lee2018simple}. 
The OOD detection problem boils down to designing $\mathcal{I}(\cdot)$ such that it assigns in-domain inputs lower scores than out-of-domain inputs.

There are two different scenarios that we consider in this work, depending on if we have any in-domain labels for data $x_i \in \mathcal{D}_{n}$. Here we define in-domain labels as any specific supervised task labels, such as sentiments, intents or topics of the text.

\paragraph{With in-domain labels} Suppose that we have multi-class label $y_i \in [K]$ and $\mathcal{D}_{n} = \{(\pmb{x}_i, y_i)\}_{i=1}^n$. Given a classifier $h$ trained with $\mathcal{D}_n$, we can use maximum calibrated softmax probability with temperature scaling as the anomaly score \citep{liang2018enhancing,hinton2015distilling}:

\begin{equation*}
     \mathcal{I}(\pmb{x}):=  - \max_{i \in [K]} \frac{\exp\left(h_i(\pmb{x})/ T\right)}{\sum_{j=1}^{K}\exp\left(h_j(\pmb{x})/ T\right)}\,, 
\end{equation*}
 where $h_i(\pmb{x})$ is the output logits of the multi-class classifier, 
 and $T$ is the temperature that is selected such that the true positive rate is at a given rate (e.g., $95\%$ in \citet{liang2018enhancing}). This method is known as {Maximum Softmax Probability (MSP)}, which requires multi-class labels to train a classifier and thus limits its application in practice. We argue that requiring in-domain labels is a less practical scenario for OOD detection and will not be further discussed it in this paper.

\paragraph{Without in-domain labels} The setting of no in-domain labels is our major focus. 
Under this assumptin, the models we can obtain in hand are usually not classifiers, but feature extractors instead. Then it is natural to resort to classic outlier detection methods like one-class support vector machine \citep{scholkopf2001estimating}, support vector data description \citep{tax2004support} or kernel density estimation (KDE) for estimating the support or the density of the in-domain data distribution.

When applying such methods to text data, the major focus of prior work is to design a good network structure or learning objectives~\citep{ruff2018deep}.
Instead, in this paper we mainly focus on how to obtain good representations from pre-trained transformers, while still obtaining good OOD detection performance.

\section{Model and Feature Learning}
\label{sec:pretraining_models}

BERT and its variants such as RoBERTa \citep[e.g.,][]{devlin-etal-2019-bert, liu2019roberta} are pre-trained on large-scale public data (denoted as $\mathcal{D}_{\rm{pub}}$) using self-supervised tasks, such as language model and next sentence prediction. These models show promising results when transferred to tasks in other domains.  We aim to leverage features obtained from pre-trained transformers to construct OOD detectors in lieu of in-domain labels in $\mathcal{D}_{n}$.
 
\subsection{BERT features for OOD detection}
After pretraining, we can obtain a BERT/RoBERTa model $f$ with $L$ layers.
We denote $f_{\ell}(\bm{x}) \in \mathbb{R}^{d}$ as the $d$-dimensional feature embeddings corresponding to the $\ell$-th layer for input $\bm{x}$, and $f(\pmb{x})$ is the overall representation using all layers of $f$. 
We explore the following methods to extract BERT features, on which we can directly construct an OOD detector using existing pure sample-based methods, such as a one-class support vector machine (OC-SVM).\footnote{It is also possible to use other related one-class classification methods, such as Isolation Forest. However, in practice we find OC-SVM works the best and 
we use it in our empirical evaluations.}

\paragraph{Features from the $\ell$-th layer $f_{\ell}$}
Options to extract $f_{\ell}(\pmb{x})$ include using the hidden states of [CLS] token or averaging all contextualized token embeddings at the $\ell$-layer. 

\paragraph{Features from all layers} Using BERT features from only one layer might not be sufficient, as prior work \citep{Jawahar2019WhatDB} has explored that different layers of BERT capture distinct linguistic properties, e.g., lower-level features capturing lexical properties, middle layers representing syntactic properties, and semantic properties surfacing in higher layers. The effects of BERT features from different layers on detecting OOD data are yet to be investigated.
One straightforward way that leverages all $L$ layers 
is to concatenate all layer-wise features $f_{\ell}(\pmb{x})$, which has no information loss. However, this solution is computationally expensive and thus hard to optimize OC-SVM or kernel based methods. 
Another solution is to perform aggregation like max- or mean-pooling along the feature dimension across all layers, sacrificing some information for efficiency. 

In this paper, we propose a simple yet effective method (described below) to use latent representations from all layers of a pre-trained transformer and can automatically decide which layers' features are important. 
This method is computationally efficient, only requiring us to solve a low-dimensional constrained convex optimization.

\paragraph{Mahalanobis distance as features (MDF) for all layers}

Support Vector Data Description (SVDD)~\citep{tax2004support} is a technique related to OC-SVM where a hypersphere is used to separate the data instead of a hyperplane.
However, the features provided by deep models may not be separable by hyperspheres.
We focus on a generalization of the hypersphere called hyper-ellipsoid to account for such surface shapes.

Suppose that we use the concatenated features from all layers $\Phi(\pmb{x}) = [f_{1}(\pmb{x}), \ldots, f_{L}(\pmb{x})]^{\top} \in \mathbb{R}^{d\cdot L}$
and consider the following optimization problem to find the hyper-ellipsoid, which is similar to the optimization formula of SVDD:
\scalebox{.9}{\parbox{.5\textwidth}{
\begin{align}
    &\min_{R, \bm{c}, \bm{\Sigma}, \bm{\xi}} ~~~~~~R^2 + \frac{1}{\nu n}\sum_i \xi_i\,, \notag \\
    &\text{s.t.}~ \|\Phi(\bm{x}_i) - \bm{c}\|_{\bm{\Sigma}^{-1}}^2 \leq R^2 + \xi_i\,,~ \xi_i \geq 0\,,\forall i\,,
\label{eq:objective}
\end{align}
}}
where $\Phi$ is the feature map,  $\bm{c}$ is the center of the hyper-ellipsoid, and $\bm{\Sigma}$ is a symmetric positive definite matrix that reflects the shape of the ellipsoid. $\xi$ is the slack variable that allows a soft boundary. 
And $R$ reflects the volume of the hyper-ellipsoid. %\footnote{We can further assume $\|\bm{\Sigma}\| =1$, where the norm can be the operator norm or Frobenius norm, which can give the definition of the hyper-ellipsoid with unique $\bm{\Sigma}$ and $R$.}
The hyper-parameter $\nu \in (0, 1]$ controls the trade-off between $R$ and $\xi$ in the objective.

%Here we also introduce a regularization term $\frac{1}{2} \|\bm{\Sigma}\|_{\text{Fr}}^2$ to constrain the complexity of $\bm{\Sigma}$. If $\bm{\Sigma} = \bm{\mathrm{I}}$, then the optimization problem is identical to one-class SVDD. 

Solving Eq~\eqref{eq:objective} exactly can be difficult, since it involves finding the optimal $\bm{\Sigma}$ of shape $D \times D$, where $D = d\cdot L$ is the dimension of the features. 
For the concatenated features $\Phi(\bm{x})$, 
$D$ can be tens of thousands or even hundreds of thousands, 
which makes the exact solution computationally intractable.
To tackle the problem, we consider a \textbf{simple and computationally efficient} approximation of the solution, which can be useful in practice.

First, we decompose the feature space into several subspaces,
based on the features from different layers, i.e., assume $\bm{\Sigma}$ is a block diagonal matrix, and
% \scalebox{.8}{\parbox{.5\textwidth}{
% \begin{align*}
%     \bm{\Sigma}=\left[
% \begin{array}{ccccc}
% \bm{\Sigma}_1& & & & \\
% &\bm{\Sigma}_2& & & \\
% & & &\ddots&  \\
% & & & &\bm{\Sigma}_L\\
% \end{array}\right]\,,
% \end{align*}
% }}
$\bm{\Sigma}_\ell$ reflects the shape of feature distribution at layer $\ell$. By a straightforward calculation, we have:
\begin{align*}
    \|\Phi(\bm{x}) - \bm{c}\|_{{\bm{\Sigma}}^{-1}}^2 = \sum_{\ell=1}^L \|f_{\ell}(\bm{x}) - \bm{c}_\ell\|_{\bm{\Sigma}_\ell^{-1}}^2 \,,
\end{align*}
where we decompose the center $\bm{c}$ to be the center of each layer $\bm{c} = [\bm{c}_1, \ldots, \bm{c}_L]^{\top}$.
Still, optimizing $\bm{c}_{\ell}$ and $\bm{\Sigma}_{\ell}$ can be difficult since the dimension of $f_{\ell}(\bm{x})$ can be high. 
Based on the intuition that $\bm{c}_\ell$ and $\bm{\Sigma}_\ell$ should not deviate from the empirical mean and covariance estimation $\widehat{\bm{c}}_{\ell}$ and $\widehat{\bm{\Sigma}}_{\ell}$ from the training data, we can replace $\bm{c}$ and $\bm{\Sigma}_{\ell}$ with the following approximation:
\scalebox{.9}{\parbox{.5\textwidth}{
\begin{align*}
        \bm{c}_\ell \approx & \widehat{\bm{c}}_\ell = \frac{1}{n} \sum_{i=1}^n \left[ f_{\ell}(\bm{x}_i) \right]\,,\\ 
    \bm{\Sigma}_\ell \approx & \frac{\widehat{\bm{\Sigma}}_\ell}{w_\ell}  = \frac{1}{(n-1)w_\ell} \sum_{i=1}^n (f_{\ell}(\bm{x}_i) - \widehat{\bm{c}}_\ell)(f_{\ell}(\bm{x}_i) - \widehat{\bm{c}}_\ell)^\top,
\end{align*}
}}
where $w_\ell$ is a layer-dependent constant. Now we only need to find proper $\{w_\ell\}_{\ell=1}^{L}$ as well as the corresponding $R$ and $\{\xi_i\}_{i=1}^n$, which is a low-dimension optimization problem that only scales linearly with the number of layer $L$. We further define:
\begin{align*}
    M_\ell(\bm{x}_i) = ( f_{\ell}(\bm{x}_i) - \widehat{\bm{c}}_\ell)^\top \widehat{\bm{\Sigma}}_{\ell}^{-1}(f_{\ell}(\bm{x}_i) - \widehat{\bm{c}}_\ell)\,,
\end{align*}
where the square root of $M_{\ell}(\pmb{x}_i)$ is also referred to as the \textit{Mahalanobis distance} of the features of data $\bm{x}_i$ from layer $\ell$.
Assume $\bm{w} = [w_1, \ldots, w_L]^\top \in \mathbb{R}^{L}$ and $M(\bm{x}) = [M_1(\bm{x}), \ldots, M_L(\bm{x})]^\top \in \mathbb{R}^{L}$, then we have:
\begin{align*}
    \|\Phi(\bm{x}) - \bm{c}\|_{\bm{\Sigma}^{-1}}^2 = \langle \bm{w}, M(\bm{x})\rangle\,.
\end{align*}
%As $\|\bm{\Sigma}\|_{\text{Fr}}^2 = \sum_{\ell=1}^L \frac{\|\widehat{\bm{\Sigma}}_l\|_{\text{Fr}}^2}{w_\ell^2}$ is not convex w.r.t $\bm{w}$, we instead minimize $-\frac{1}{2} \|\bm{w}\|_2^2$, which has a similar regularization effect on $\bm{\Sigma}$ (as we don't want $\|\bm{w}\|_2$ to be small, which can make $\|\bm{\Sigma}\|_{\text{Fr}}$ very large). 
We additionally add a $\ell_2$ regularization term to $\bm{w}$ as we don't want it to be large.
So the final optimization problem to solve is:
\begin{align}
    & \min_{R, \bm{w}, \bm{\xi}} \frac{1}{2}\|\bm{w}\|_2^2 + R^2 + \frac{1}{\nu n} \sum_i \xi_i, \notag\\
    & \quad \text{s.t.}\quad \langle \bm{w}, M(\bm{x}_i) \rangle \leq R^2 + \xi_i,\ \xi_i \geq 0\,, \forall i\,, \label{equ:our_obj}
\end{align}
which in fact is  a one-class SVM with a linear kernel, with Mahalanobis distance of each layers as features (MDF), and it can be solved with the standard convex optimization. 
We illustrate our proposed algorithm in Figure \ref{fig:overview}.

\paragraph{Remark} Note that the optimization in Eq~\eqref{equ:our_obj} is not identical as that in Eq~\eqref{eq:objective}, since we are using empirical 
sample mean $\{\widehat{\bm{c}}_{\ell}\}_{\ell=1}^{L}$ and covariance $\{\widehat{\bm{\Sigma} }_{\ell} / w_{\ell}\}_{\ell=1}^{L}$ to 
replace the original parameters $\bm{c}$ and $\bm{\Sigma}$ in Eq~\eqref{eq:objective}, which are hard to optimize when the dimension of the concatenated features $\Phi(\bm{x})$ is high.
Also, our approximation from Eq~\eqref{eq:objective} to Eq~\eqref{equ:our_obj} is different from 
the known result that when $\Phi(\bm{x})$ is the infinite-dimensional feature map of the widely used Gaussian RBF kernels, 
OC-SVM and SVDD are equivalent and asymptotically 
consistent density estimators
 \citep{tsybakov1997nonparametric,vert2006consistency}.
In our case, $\Phi(\bm{x})$ is the concatenated features from all layers of pre-trained transformers, 
which makes our approximation fundamentally different from prior work.

\begin{table*}[t]
    \centering
    \scalebox{0.915}{
    \begin{tabular}{ l l l }
    \toprule
  \multicolumn{3}{c}{Cross-corpus Examples (SST)} \\ \hline
    \textbf{Type~~~~~~~~~} & \textbf{Source~~~~~~~~~~} & \textbf{Text}~~~~~~~~~~~~~~~~~~~~~~\\
        \hline
      In-Domain & SST & \emph{if you love reading and or poetry , then by all means check it out}\\
    %   In-Domain & SST & \emph{Much of it comes from the brave , uninhibited performances by its lead actors .}\\
      In-Domain & SST & \emph{there 's no disguising this as one of the worst films of the summer}\\
    \hline
  Out-of-Domain & RTE &\textit{capital punishment is a deterrent to crime}\\
  Out-of-Domain & SNLI & \textit{a crowd of people are sitting in seats in a sports ground bleachers }\\
  Out-of-Domain & Multi30K & \textit{a trailer drives down a red brick road}\\
    \bottomrule
    \multicolumn{3}{c}{Cross-intent Examples (CLINIC150)} \\ \hline
    \textbf{Type~~~~~~~~~} & \textbf{Intent~~~~~~~~~~} & \textbf{Text}~~~~~~~~~~~~~~~~~~~~~~\\
        \hline
      In-Domain & Transfer & \emph{move 100 dollars from my savings to my checking}\\
      In-Domain & PTO Request & \emph{let me know how to make a vacation request}\\
      % In-Domain & Change Language & \textit{switch the language setting over to german}\\
      % In-Domain & Distance & \emph{tell the miles it will take to get to las vegas from san diego}\\
      % In-Domain & Travel Suggestion & \emph{what sites are there to see when in evans}\\
      % In-Domain & Todo List Update & \textit{nuke all items on my todo list}\\
      % In-Domain & Text & \emph{send a text to mom saying i'm on my way}\\
      In-Domain & Food Last & \emph{is rice ok after 3 days in the refrigerator}\\
    %   In-Domain & Tell Joke & \emph{can you tell me a joke about politicians}\\
    %   In-Domain & Rewards Balance & \emph{how high are the rewards on my discover card}\\
    \hline
  Out-of-Domain & --- &\textit{how are my sports teams doing}\\
%   Out-of-Domain & --- & \textit{create a contact labeled mom}\\
  Out-of-Domain & --- & \textit{what's the extended zipcode for my address}\\ \bottomrule
\end{tabular}
}
\vspace{-.3em}
\caption{Examples of in-domain/out-of-domain samples for SST and CLINIC150. The source labels for SST and the intent labels for CLINIC150 are here just for illustration and are not included in $\mathcal{D}_n$. None of the above OOD samples are provided in training as well.}
\label{tab:example_sentences}
\vspace{-0.5em}
\end{table*}

\subsection{Feature fine-tuning}
We can also fine-tune the pre-trained transformer $f$ on the \textbf{unsupervised} in-domain dataset $\mathcal{D}_{n}$ so that $f(\pmb{x})$ can better represent the distribution of $\mathcal{D}_n$. We explore two domain-specific fine-tuning approaches.

\paragraph{In-domain masked language modeling (IMLM)}
\citet{gururangan2020don} find that domain-adaptive masked language modeling~\citep{devlin-etal-2019-bert} would improve supervised classification capability of BERT when it is transferred to that domain. 
Similarly, we can do MLM on $\mathcal{D}_n$ and argue this would make the features of $\mathcal{D}_n$ concentrate, bringing benefits to downstream OOD detection.

\paragraph{Binary classification with auxiliary dataset (BCAD)}
Another way of fine-tuning the model $f$ is to use the public dataset $\mathcal{D}_{\rm{pub}}$ that pretrains it. We consider the training data in $\mathcal{D}_{n}$ as in-domain positive samples and data in the public dataset $\mathcal{D}_{\rm{pub}}$ as OOD negative samples. We add a new classification layer on top of $f$ and update this layer together with all parameters of $f$ by performing a binary classification task. In practice, we only need a small subset of $\mathcal{D}_{\rm{pub}}$, denoted as $\Tilde{\mathcal{D}}_{\rm{pub}}$,  for fine-tuning.
Since $\Tilde{\mathcal{D}}_{\rm{pub}}$ is publicly available and has no labels, we do not violate the unsupervised setting. $\Tilde{\mathcal{D}}_{\rm{pub}}$ does not provide any information about the OOD samples at test time as well. 

Besides, the added classification layer can actually be applied for OOD detection using the MSP method, and this is exactly the setting of \textit{zero-shot} classification, which we use as a baseline for comparison in our experiments.

\section{Experiments}
\label{sec:exp}
\paragraph{Datasets}
We consider two distinct datasets for experiments, where one is to regard text from unseen corpora as OOD, and the other one is to detect class-level OOD samples within the same corpus.

\myparagraph{$\bullet$~~
Cross-corpus dataset (SST)}~We follow the experimental setting in~\citet{hendrycks2020pretrained}, by providing in-domain $\mathcal{D}_n$ with the original training set of SST dataset~\citep{Socher2013RecursiveDM} 
and considering samples from four other datasets (i.e., 20 Newsgroups~\citep{Lang1995NewsWeederLT}, English-German Multi30K~\citep{Elliott2016Multi30KME}, RTE~\cite{Dagan2005ThePR} and SNLI~\citep{Bowman2015ALA}) as OOD data.
For evaluation, we use the original test data of SST as in-domain positives and randomly pick 500 samples from each of the four datasets as OOD negatives.
We do not include any sentiment labels from SST to $\mathcal{D}_n$ for training. %Besides, we do lowercase and remove some special tokens to make their consistent in format. 

\myparagraph{$\bullet$~~
Cross-intent dataset (CLINIC150)}~This is a crowdsourced dialog dataset~\citep{larson2019eval}, including in-domain queries covering 150 intents and out-of-domain queries that do not fall within any of the 150 intents.
We use all 15,000 queries that are originally in its training data as in-domain samples but discard their intent labels. For evaluation, we mix the 4,500 unseen in-domain test queries with 1,000 out-of-domain queries and wish to separate two sets by their anomaly scores. 

Examples taken from the two datasets can be found in Table~\ref{tab:example_sentences}. Note that for both datasets, only the in-domain samples are used for training, and the source/intent labels are not used in our experiments. 

%Illustrative in-domain and out-domain examples for both datasets can be found in Table \ref{tab:example_sentences}. Note for CLINIC150, the intent labels are not used in our experiments. 

\paragraph{Evaluation metrics}
We rank all test samples by their anomaly scores and follow \citet{liang2018enhancing} to report four different metrics, namely, Area Under the Receiver Operating Characteristic Curve (\textbf{AUROC}), Detection Accuracy (\textbf{DTACC}), and Area under the Precision-Recall curve (AUPR) for in-domain and out-of-domain testing sentences respectively, denoted by \textbf{AUIN} and \textbf{AUOUT}. 

\paragraph{Model configurations}
We evaluate all methods with both BERT and RoBERTa (\texttt{base} models with 768 latent dimensions and 12 layers).

\paragraph{Choice of $\Tilde{\mathcal{D}}_{\rm{pub}}$ for BCAD} We adopt the {BooksCorpus}~\cite{Zhu2015AligningBA} and English Wikipedia, which are the sources used in common by  BERT and RoBERTa for pre-training. We split paragraphs into sentences and sample $\Tilde{\mathcal{D}}_{\rm{pub}}$ to have the same size as $\mathcal{D}_n$ for BCAD. 

\paragraph{Baselines}
To examine the effectiveness of our newly proposed anomaly score based on MDF that utilizes the representations of all layers, we compare it with the following baselines.

    \vspace{0.5em}
    
    $\bullet$ (Ro)BERT(a)-Single layer: It uses $f_{\ell}(\pmb{x})$ mentioned above.
    We iterate all 12 layers and detailed results of each layer are discussed in Section~\ref{sec:single_layer}. 

    \vspace{0.5em}
    
    $\bullet$ (Ro)BERT(a)-Mean pooling: we construct all-layer representation by averaging all $f_\ell(\pmb{x})$, which has  768 dimensions. 
    
    \vspace{0.5em}
    
    $\bullet$ (Ro)BERT(a)-Max pooling: we aggregate all layers by picking largest values along each feature dimension and get a 768-dimension vector.
    
    \vspace{0.5em}
    
    $\bullet$ (Ro)BERT(a)-Euclidean distance as features (EDF): we replace Mahalanobis distance with Euclidean distance and still obtain a 12-dimension vector.
    
    \vspace{0.5em}
    
    $\bullet$  TF-IDF: we extract TF-IDF features and adopt SVD to reduce high-dimensional features to 100 dimensions for computational efficiency. 
    
     All of the above methods extract features as the input to OC-SVM to compute anomaly scores. 
     
    \vspace{0.5em}
    
    $\bullet$  BCAD + MSP:  It performs zero-shot classification after BCAD fine-tuning, as discussed in Section~\ref{sec:pretraining_models}. The temperature scaling is tuned to achieve the best result. This method is not applicable when no $\Tilde{\mathcal{D}}_{\rm{pub}}$ is provided.

\begin{table}[t]
\centering
 \setlength{\tabcolsep}{2pt}
		\renewcommand{\arraystretch}{1.1}
\scalebox{0.92}{
\begin{tabular}{lcccc|cccc}
\toprule
 {Layer} & \multicolumn{4}{c}{SST} & \multicolumn{4}{c}{CLINIC150} \\ \hline
 & \multicolumn{2}{c}{BERT}  & \multicolumn{2}{c}{RoBERTa} & \multicolumn{2}{c}{BERT} & \multicolumn{2}{c}{RoBERTa} \\ \hline
    & ~\small CLS & ~\small AVG & ~ \small CLS &~ \small AVG  &~  \small CLS & ~\small AVG &~  \small CLS&~ \small AVG\\ \hline
12 & \textbf{92.7} & \textbf{81.7} & \textbf{89.8} & \textbf{87.8}  & 61.5 & 60.2 & 53.4 & 51.6 \\
11 & 88.8 & 66.3 & 88.8 & 68.8  & 57.3 & 59.0 & 51.6 & 55.5 \\
10 & 87.7 & 52.1 & 79.6 & 68.4  & 56.6 & 55.4 & 53.8 & 56.2 \\
9 & 85.5 & 50.7 & 84.2 & 67.2  & 56.8 & 56.5 & 58.3 & 56.5 \\
8 & 82.9 & 57.6 & 78.7 & 67.7  & 61.6 & 55.8 & \textbf{58.9} & 56.0 \\
7 & 85.8 & 59.2 & 83.6 & 67.5  & \textbf{62.3} & 63.0 & 57.5 & 56.4 \\
6 & 76.4 & 61.9 & 73.0 & 67.8  & 58.2 & 62.3 & 55.5 & 56.7 \\
5 & 74.2 & 58.2 & 63.5 & 67.2  & 56.3 & 62.8 & 56.2 & 57.1 \\
4 & 66.7 & 67.4 & 70.0 & 69.8   & 61.9 & 60.9 & 52.7 & 57.8 \\
3 & 65.8 & 67.5 & 62.9 & 69.3  & 54.3 & 59.4 & 51.0 & 58.5 \\
2  & 62.6 & 63.2 & 75.7 & 68.8  & 60.4 & 58.6 & 55.6 & \textbf{59.9} \\
1 & 68.1 & 63.5 & 70.0 & 71.0  & 60.9 & \textbf{64.6} & 55.6 & 58.5 \\
\bottomrule
\end{tabular}
}
\caption{The \textbf{AUROC} scores of OOD detection on the SST/CLINIC150 dataset for each layer of BERT/RoBERTa. CLS denotes using the hidden state of the [CLS] token and AVG represents averaging all token embeddings in the same layer. Layer 12 indicates the top layer and layer 1 is the bottom layer right after the word embedding layer. The best result for each column is marked in bold. }
\label{tab:probing}
\end{table}

\section{Results and Discussions}
In this section, we present the results for our experiments and summarize our findings.

\begin{table*}[t]
 \setlength{\tabcolsep}{4pt}
		\renewcommand{\arraystretch}{1.02}
		\centering
\scalebox{.91}{
\begin{tabular}{l|c|cccc|cccc}
\toprule
 &  & \multicolumn{4}{c}{SST} & \multicolumn{4}{c}{CLINIC150} \\ 
& \small\#feats & \footnotesize AUROC  & \footnotesize DTACC & \footnotesize AUIN & \footnotesize AUOUT & \footnotesize AUROC &  \footnotesize DTACC & \footnotesize AUIN & \footnotesize AUOUT \\ \midrule\midrule 
BERT-Single layer (best)            & 768      & 92.7  & 85.8 & 93.4 & 91.7  & 64.6 & 60.9 & 88.4 & 26.7  \\
RoBERTa-Single layer (best)        & 768      &  89.8   &  91.5  & 79.2 & 93.8 & 59.9 & 57.6 & 86.8 & 22.7 \\
\midrule\midrule
BERT +  Mean-Pooling     & 768      &  81.8 & 76.5 & 77.2 & 82.8 & 62.9 & 59.9 & 87.0 & 27.9 \\
BERT +  Max-Pooling       & 768      &  67.2 & 66.1 & 64.2 & 59.4 & 63.0 & 60.0 & 88.0 & 25.8  \\
RoBERTa +  Mean-Pooling            & 768      &  91.0 & 92.3 & 80.9 & 94.5 & 57.1 & 56.2 & 85.7 & 20.5 \\
RoBERTa +  Max-Pooling            & 768      &  93.2 & 91.9 & 89.3 & 95.1 & 54.9 & 54.4 &  84.8  & 19.4 \\
\midrule\midrule
BERT + EDF  & 12	& 90.1   &	84.8 &	92.8 & 84.2  &  55.3 & 55.2  & 84.3 &  20.3 \\
BERT +  MDF            & 12      & 93.3	& 87.5   &	94.9 &	89.1 & 76.7 & 71.1 &  93.4 & 38.2 \\
BERT + IMLM +  MDF      & 12       & 93.6 &	88.1 &	97.5 &	89.4 & 77.8 & 72.2 & 93.8 & 39.1    \\
BERT + BCAD +  MDF    & 12    & 97.0  &	94.5	& 98.0 &	94.8  & 81.2  &	74.5  &	94.6  &	47.4   \\
BERT + IMLM + BCAD +  MDF & 12 & 98.1 &	95.4 &	98.7 &	95.9  & 82.1  &	75.6  &	95.0  &	47.6   \\ \hline
RoBERTa + EDF  & 12 & 99.5 &  95.8 &  99.5 & 99.4 & 56.9 &  56.9 & 86.3 & 19.6 \\
RoBERTa +  MDF      & 12   & 99.8  & 97.7 & 99.8  & 99.8 & 78.6 & 71.9 & 93.8 & 42.6  \\
RoBERTa + IMLM +  MDF   & 12  &  99.9  & 97.8 & 99.8 & 99.8  & 80.1 & 73.1 & 94.5  & 44.9      \\
RoBERTa + BCAD +  MDF   & 12  & 99.2 & 96.6 &  99.4 & 98.7 & 80.5 & 72.9 & 94.3 & 49.4   \\
RoBERTa + IMLM + BCAD +  MDF & 12 & \textbf{99.9} & \textbf{98.6} &  \textbf{99.9} &\textbf{99.9} & \textbf{84.4}	 & \textbf{76.7}	 & \textbf{95.4}	 & \textbf{59.9} \\
\midrule\midrule
TF-IDF + SVD             & 100      &   78.0 & 72.0 & 78.2 & 73.2   & 58.5  &  56.5  &  86.2    &21.8 \\
BERT + BCAD + MSP        & - & 68.5  &	69.0  &	61.5  &	65.4 & 68.3	 & 63.5	 & 89.7	 & 34.1 \\
RoBERTa + BCAD + MSP    & -    & 73.7  & 69.3  & 69.0  & 75.3 & 62.1 & 59.6  & 85.9 &  27.8     \\
\bottomrule
\end{tabular}
}
\caption{OOD detection performance on SST and CLINIC 150 for all models. OC-SVM is used for computing anomaly scores except MSP, and its parameters size is \#feats. For (Ro)BERT(a)+Single-layer, the best results in Table~\ref{tab:probing} are reported. For all MDF-based model, we only report results of AVG as sequence representation at each layer due to space limit. Larger values of all four metrics indicate better performances. The best result for each metric is marked in bold. }
\label{tab:results}
\end{table*}

\subsection{Using single-layer feature $f_{\ell}(x)$}
\label{sec:single_layer}

%%%%%%%%% 
Table~\ref{tab:probing} shows results obtained from using the [CLS] embedding or averaging token embeddings (AVG) at each layer of (Ro)BERT(a) models in the cross-corpus and the cross-intent dataset. 

We observe that detecting cross-intent OOD samples in CLINIC150 is more challenging than that of cross-dataset OOD data in SST. This is mainly because the OOD samples in CLINIC150 are sorted by humans and the differences between intents can be subtle. We will further compare the performance of these two settings in Figure~\ref{fig:aurocs}. 

\myparagraph{The best $f_{\ell}(\pmb{x})$ for OOD is dataset-specific} For the cross-corpus dataset (SST), we find that the best results come from the top layer of both (Ro)BERT(a). However, for the cross-intent dataset (CLINIC150), the middle layers perform the best when using [CLS], while the bottom layers achieve the best results with AVG. This indicates that OOD distributions are not simply based on certain types of linguistic features and the strategy of choosing $f_{\ell}(\pmb{x})$ is dataset-specific; for some dataset, semantic features play a more important role, while sometimes we need to focus on syntactic or lexical features. This validates the assumption that it is beneficial to fully utilize all layers of the hidden representations from pre-trained transformers to detect OOD instances. 

We find using $f_{\ell}{(\pmb{x})}$ of BERT is generally better than RoBERTa, especially with [CLS]. We guess next sentence prediction may cause this, which pre-trains on [CLS] and is exclusive for BERT. 

In later sections, (Ro)BERT(a)-Single layer will refer to the best one in Table~\ref{tab:probing}.

\newcommand{\gapp}{-0.5em}

\begin{figure*}[t]
    \centering
     \begin{tabular}{cccc}
        \includegraphics[height=0.205\linewidth]{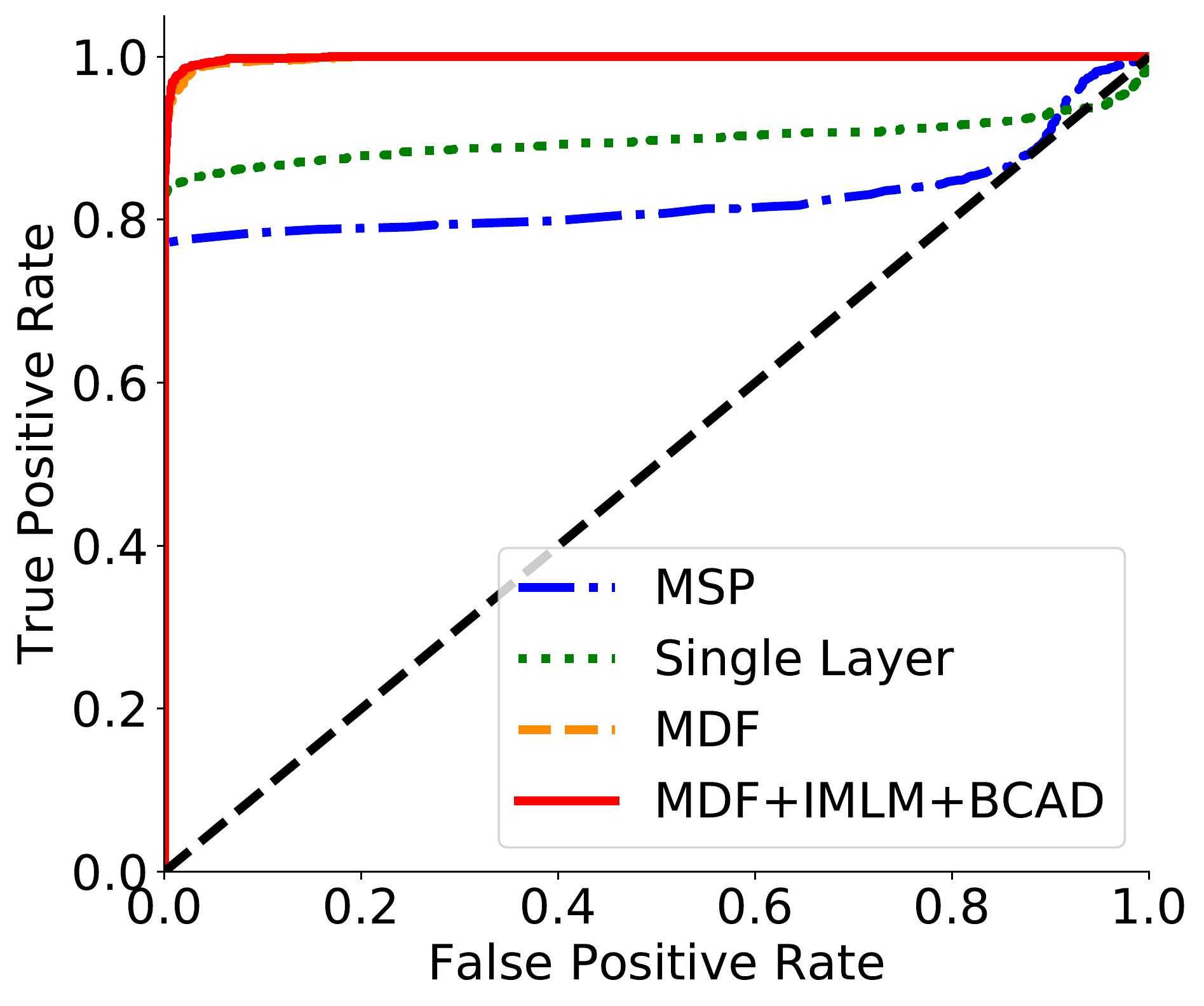} & 
         \hspace{\gapp}
         \includegraphics[height=0.205\linewidth]{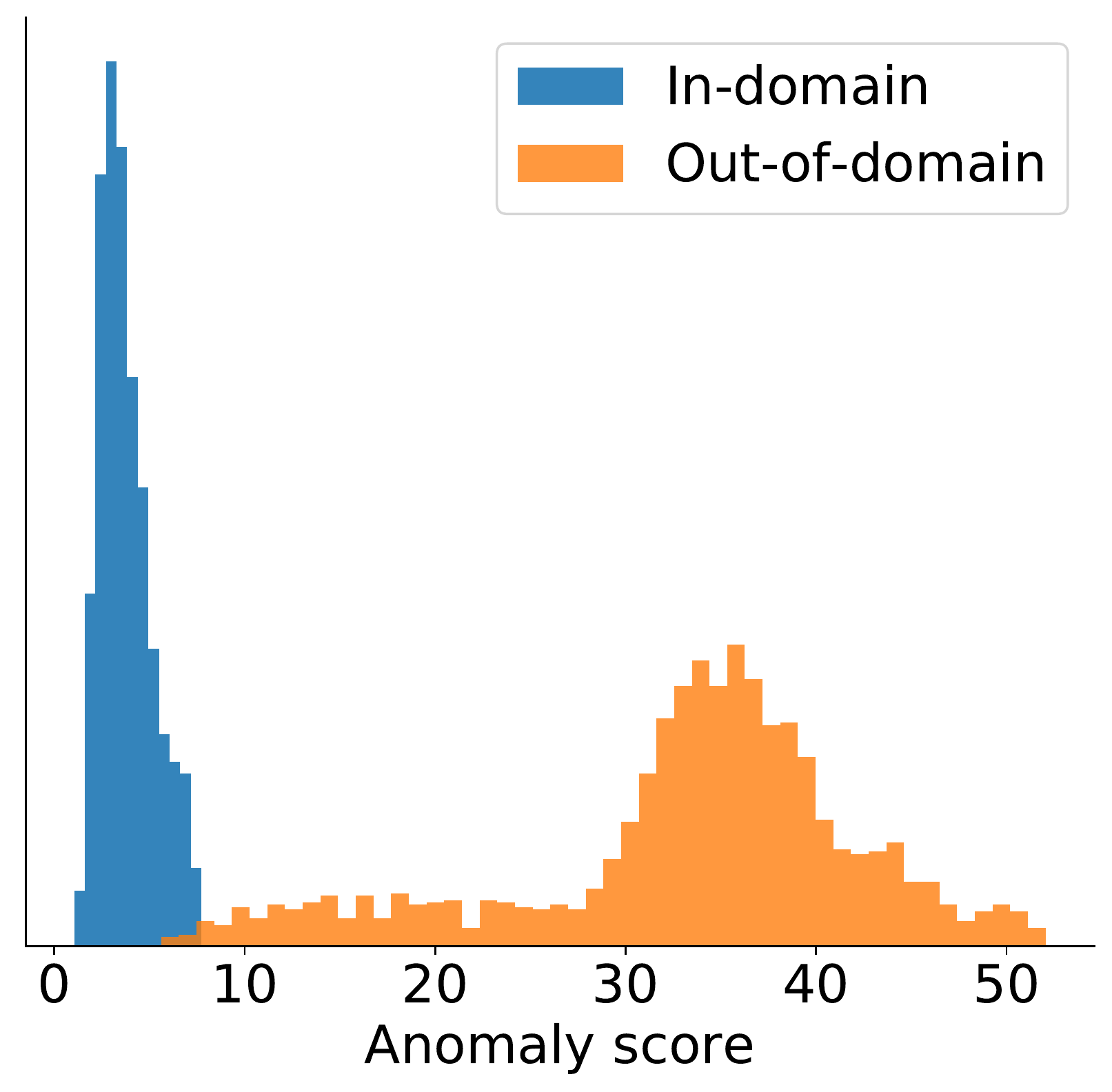} &
         
          \includegraphics[height=0.205\linewidth]{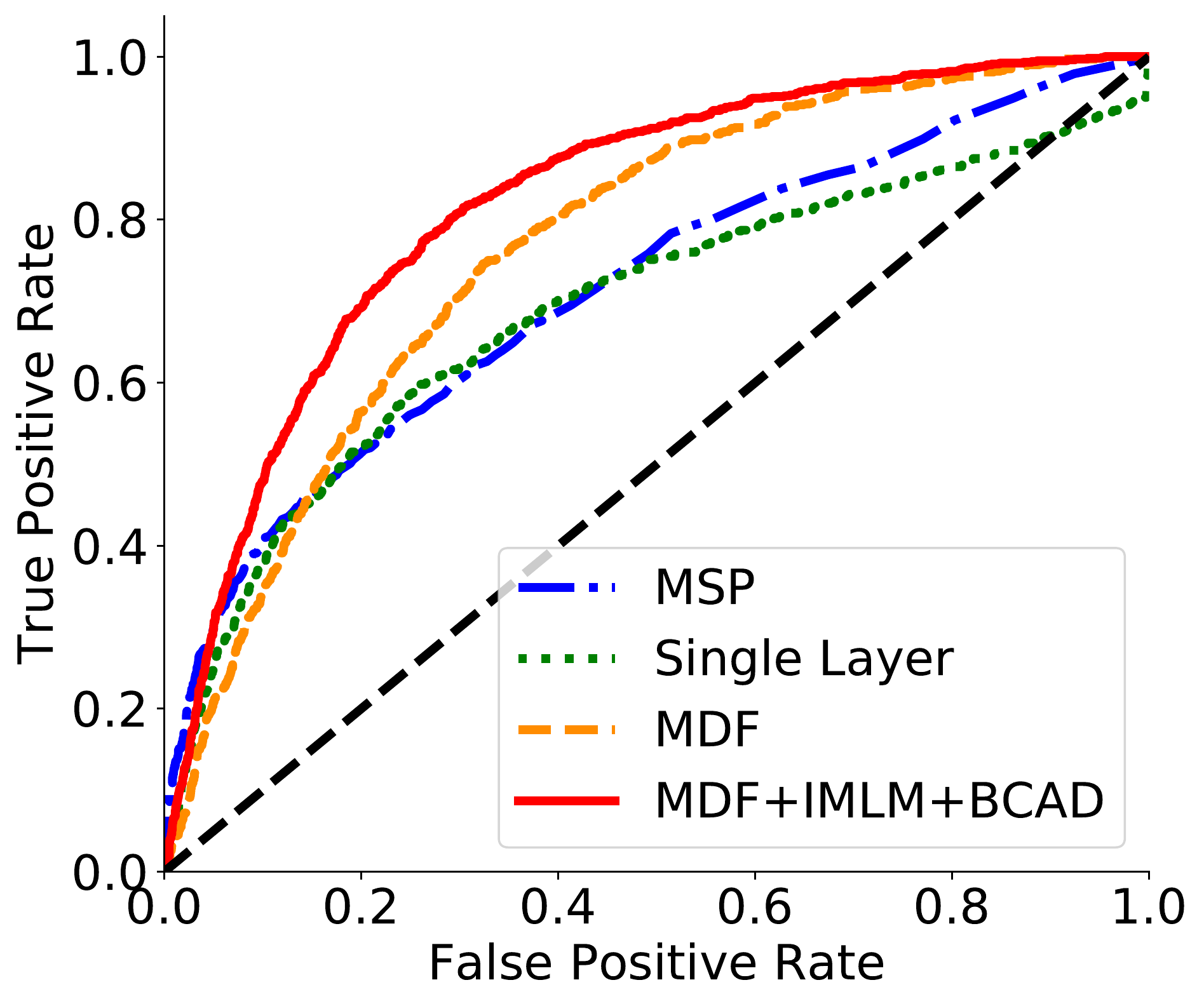}& 
          \hspace{\gapp}
         \includegraphics[height=0.205\linewidth]{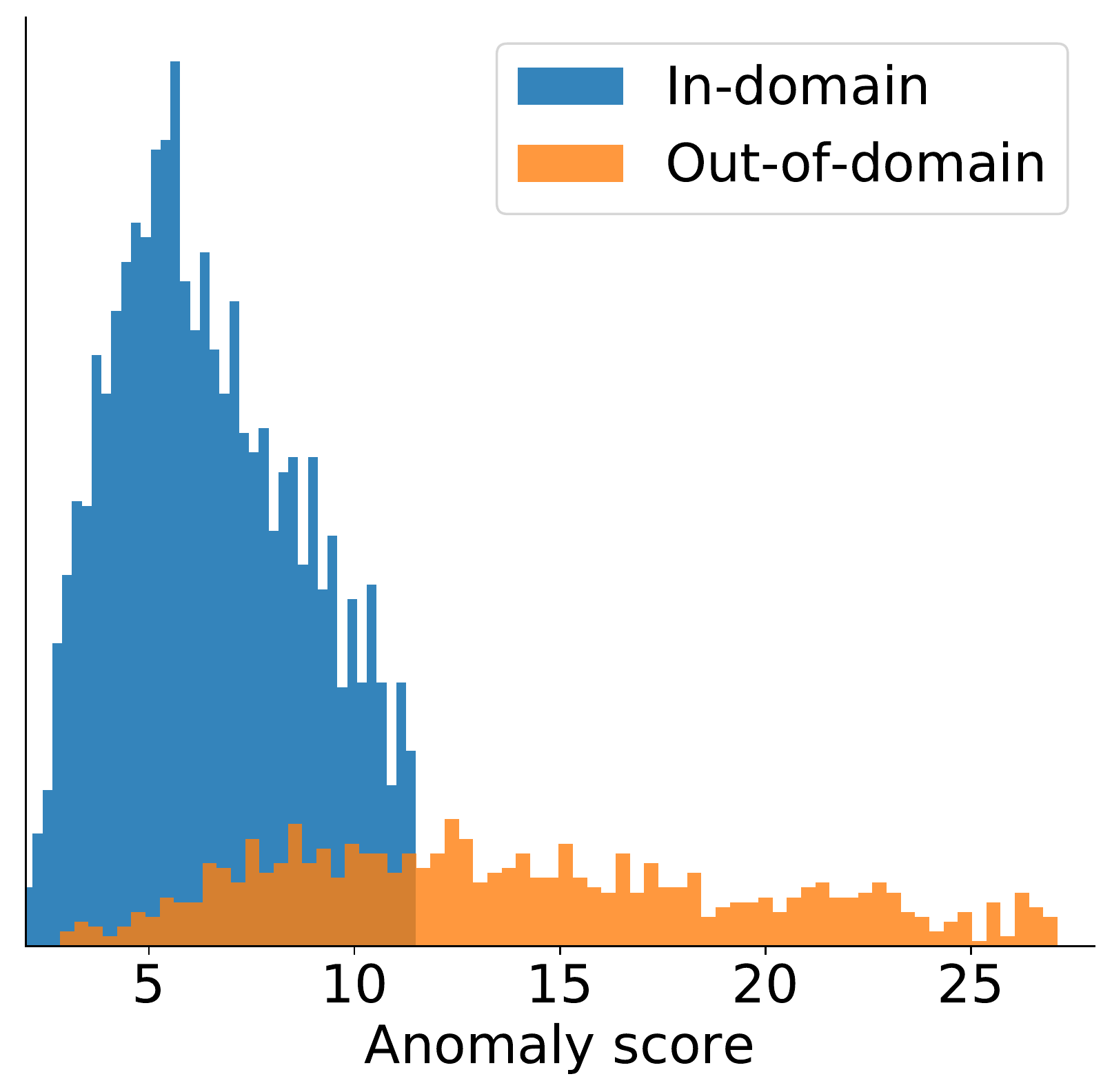}
         \vspace{-.5em}\\
         
          \footnotesize{(a)  ROC Curve on SST} &
         \hspace{\gapp}
        \footnotesize{(b) $\mathcal{I}(\bm{x})$ on SST}  &
        \footnotesize{(c) ROC Curve on CLINIC150 } &
         \hspace{\gapp}
        \footnotesize{(d) $\mathcal{I}(\bm{x})$ on CLINIC150} \\
    \end{tabular}
    \vspace{-0.5em}
    \caption{\small{(a): ROC curves on the SST dataset. (b): Distribution of anomaly scores generated by IMLM + BCAD + MDF. Both figures are based on the BERT model. 
    (c): ROC curves on the CLINIC150 dataset. (d): Distribution of anomaly scores generated by IMLM + BCAD +  MDF.} }
    \label{fig:aurocs}
\end{figure*}

\begin{table*}[t]
\centering
\scalebox{0.88}{
\begin{tabular}{c|c|c|c|c|c}
\hline
& Sentence &  GT &  TF-IDF    & Single    & MDF    \\ \hline
(a) &  is a visa necessary for traveling to south africa & In & In  & In & In \\ \hline
(b) &  can you tell me who sells dixie paper plates & Out & In & Out & Out \\ \hline
 (c) & can you tell me how to solve simple algebraic equations with one variable  & Out & Out & In     & Out    \\ \hline
% how much has the dow changed today  & Out & In & In & In \\ \hline
(d) & what oil is best for chicken  & Out & In & In & In \\ \hline
\end{tabular}
}
\vspace{-.5em}
\caption{\small{Examples of CLINIC150 with predictions from three models, which is ``In'' when sample's anomaly score is lower than 25th percentile and ``Out'' when larger than 75th percentile. GT is the ground truth and Single stands for BERT-Single.}}
\label{tab:case}
\vspace{-.5em}
\end{table*}

\subsection{Overall OOD detection performance}

We report the empirical results of OOD detection in Table~\ref{tab:results} and the following observations.

\myparagraph{Pre-trained transformers produce good feature representations} Methods using single-layer feature $f_{\ell}$ outperforms frequency-based features (TF-IDF) and zero-shot classification (MSP), which validates the strong representation capability granted by self-supervised pre-training.

\myparagraph{Simple aggregations of all layers are not so effective} The results of max-pooling and mean-polling are not very promising. Even though we observe an absolute 0.5\% boost in SST using max-pooling, using the best single layer actually outperforms those simple aggregations in CLINIC150.

\myparagraph{MDF is more effective} MDF consistently outperforms methods that directly use features $f_{\ell}(\pmb{x})$, simple aggregations of $f_{\ell}(\pmb{x})$, or TF-IDF features on all four metrics. In terms of AUROC,  MDF outperforms the best single-layer of (Ro)BERT(a) by absolute 7.1\% on SST and 14.0\% on CLINIC150.

MDF also performs better than EDF. Note that Euclidean distance is a special case of Mahalanobis distance when the covariance is an identity matrix.  Empirically, the features generated by neural models are not invariant across all dimensions; and the comparison between MDF and EDF validates SVDD with a hyper-ellipsoid is better than a hypersphere.

\myparagraph{MDF is more efficient in training OC-SVM}
 Notice that our approach is also more computationally efficient when obtaining optimal $\pmb{w}$ and $\pmb{R}$ since the optimization is performed on a new transformed low dimensional data space ($d = 12$ is number of layers in $f$). See column $\#$feats in Table~\ref{tab:results} for detailed comparisons. 

\myparagraph{Fine-tuning techniques improve performance}
From Table~\ref{tab:results}, we can see both MILM and BCAD improve OOD detection performance when incorporated with MDF separately. The overall best detecting performance is achieved by MILM + BCAD + MDF, combining both proposed fine-tuning methods with MDF.

We also find that RoBERTa outperforms BERT when using MDF, even though features from a single layer prefers BERT in Table~\ref{tab:probing}.

\subsection{Visualizations}
We plot the ROC curves of four different anomaly scores on SST in Figure~\ref{fig:aurocs} (a) and on CLINIC150 in Figure~\ref{fig:aurocs} (c),
confirming that our proposed MDF and two fine-tuning techniques improve the ability in detecting OOD samples. 
We also present the distributions of anomaly scores $\mathcal{I}(\bm{x})$ generated by our best method in Figure~\ref{fig:aurocs}~(b) for SST and in Figure~\ref{fig:aurocs}~(d) for CLINIC150. For SST, the OOD detector can clearly separate $\mathcal{I}(\bm{x})$ of in-domain and out-domain samples, and the in-domain scores are densely concentrated on the low-score region. Although for CLINIC150, we do observe some OOD samples mixing with in-domain ones, accounting for the gap of metric scores between two datasets.

\subsection{Case Studies}
We present some examples from CLINIC150 together with their corresponding predictions by TF-IDF,
BERT-single layer and MDF methods in Table~\ref{tab:case}. 
TF-IDF predicts false positives for examples (b) and (d) because most of the words in the example test query are seen in the training set, like ``i would like you to buy me some paper plates'' (intent: order), ``i need to know how long to cook chicken for'' (intent: cooking time) and etc. 
BERT-single layer learns the syntax of ``can you tell me how to ...'', which is frequently seen in the training data, but it fails to discern that the semantic meaning is out-of-domain. For example (d), all models make the mistake, potentially associating it with the intent: recipe (``i need to find a good way to make chicken soup'' or ``what's the best way to make chicken stir fry'').

\section{Related Work} 
Out-of-domain detection is essentially an important component for trustworthy machine learning applications. 
There are two lines of work proposed to perform out-of-domain detection. 
One is to tackle the problem in specific multi-class classification tasks, where well-trained classifiers are utilized to design anomaly scores \citep[e.g.,][]{hendrycks2016baseline,liang2018enhancing,lee2018simple,card2019deep,hendrycks2020pretrained,xu2020deep}, 
Those methods can only be useful when multi-class labels are available,  
which limits their application in more general domains. Our proposed work goes beyond this limitation and can utilize large amounts of unsupervised data.

Another line of work is based on support estimation or density estimation, which assumes that the in-domain data is in specific support or from the high density region \citep{scholkopf2001estimating,tax2004support}. 
In principle, our work is closely related to this line of work.
Besides, \citet{pmlr-v48-zhai16,ruff2018deep,zong2018deep} also leverage the features of neural networks, though these methods require designing specific network structures for different data.
%and can not directly utilize the representation of pre-trained transformers. 
Our work circumvents the issues of prior work by designing a computationally efficient method that leverages the powerful representations of pre-trained transformers.

Finally, the fine-tuning techniques we use to improve the representation of data are closely related to unsupervised pre-training for transformers \citep{devlin-etal-2019-bert,yang2019xlnet}, and recently proposed contrastive learning \citep[e.g.,][]{he2020momentum,chen2020simple}.
Lately, \citet{gururangan2020don} discover that performing pre-training (MLM) on the target domain with unlabeled data can also help to improve downstream classification performance.
To the best of our knowledge, our method is the first to incorporate transformers and pre-training techniques to improve out-of-domain detection.

\section{Conclusion}
We study the problem of detecting out-of-domain samples with unsupervised in-domain data, which is a more general setting for out-of-domain detection. We propose a simple yet effective method using Mahalanobis distance as features, which significantly improves the detection ability and reduces computational cost in learning the detector. Two domain-adaptive fine-tuning techniques are further explored to boost the detection performance.

In the future, we are interested in deploying our OOD method to real-world applications, such as detecting unseen new classes for incremental few-shot learning \citep{zhang2020discriminative,xia2021incremental} or filtering OOD samples in data augmentations. 

\section*{Acknowledgments}
We would like to thank the anonymous reviewers
for their valuable feedback and comments.

% \paragraph{Ethical Considerations}
% This paper does not introduce new datasets nor include demographic or identifying characteristics in its analysis.
% All our experiments are conducted on GPU computing resources, and are manageable with a machine with eight GeForce RTX 2080. 
% We believe that our line of research benefits all stakeholders, and is contributing to AI safety by avoiding unpredictable outcomes of deep neural networks through out-of-domain detection. 

\bibliography{reference}
\bibliographystyle{acl_natbib}

\end{document}